\documentclass[11pt]{article}
\usepackage[superscript]{cite}
\usepackage{wordlike}
\usepackage{setspace}

\usepackage{hyperref}
\hypersetup{colorlinks,urlcolor=blue}
\usepackage{float}
\usepackage{pdflscape}
\usepackage{textcomp}
\usepackage{cite}
\usepackage{times}
\usepackage{latexsym}
\usepackage{hyperref}
\usepackage{booktabs}
\usepackage{amsmath}
\usepackage{threeparttable}
\usepackage{tabularx}
\usepackage{graphicx}
\usepackage{multirow}
\usepackage{longtable}
\usepackage{xcolor}
\usepackage{threeparttable}
\makeatletter
\definecolor{myblue1}{RGB}{47,84,150}
\newcommand{\printfnsymbol}[1]{%
  \textsuperscript{\@fnsymbol{#1}}%
}
\renewcommand{\section}{\@startsection%
  {section}%
  {0}%
  {0em}%
  {-\baselineskip}%
  {0.5\baselineskip}%
  {\color{myblue1}\Large\sffamily}}%
\renewcommand{\subsection}{\@startsection%
  {subsection}%
  {1}%
  {0em}%
  {-\baselineskip}%
  {0.5\baselineskip}%
  {\color{myblue1}\large\sffamily}}%
\makeatother

\renewcommand{\frame}{}

\title{\singlespacing DeepSeeNet: A Deep Learning Model for
Automated Classification of Patient-based
Age-related Macular Degeneration Severity
from Color Fundus Photographs}

\author{
\small
Yifan Peng, PhD$^{1,*}$, Shazia Dharssi$^{1, 2,*}$, Qingyu Chen, PhD$^{1}$, Tiarnan D. Keenan, BM BCh, PhD$^{2}$, Elvira Agr\'{o}n, MA$^{2}$, Wai T. Wong, MD$^{2}$, Emily Y. Chew, MD$^{2}$, Zhiyong Lu, PhD$^{1}$\\
1. National Center for Biotechnology Information (NCBI), National Library of Medicine (NLM), National Institutes of Health (NIH), Bethesda, Maryland, United States;\\
2. National Eye Institute (NEI), National Institutes of Health (NIH), Bethesda, Maryland, United States;\\ 
* These authors contributed equally to this work.\\
}

\date{}

\begin{document}

\maketitle

\parindent=0em
\textbf{Taxonomy topics (2-6)}

deep learning; age-related macular degeneration (AMD); Age-Related Eye Disease Study (AREDS); convolutional neural network (CNN); artificial intelligence (AI)
\vspace{1em}

\textbf{Corresponding Author(s)}

Zhiyong Lu, PhD, National Center for Biotechnology Information (NCBI), National Library of Medicine (NLM), National Institutes of Health (NIH), 8600 Rockville Pike, Bethesda, MD 20894, \url{zhiyong.lu@nih.gov}

Emily Y. Chew, MD, National Eye Institute (NEI), National Institutes of Health (NIH), 9000 Rockville Pike, Bethesda, MD 20894, \url{echew@nei.nih.gov}
\vspace{1em}

\textbf{Financial Support}: 
Supported by the intramural program funds and contracts from the National
Center for Biotechnology Information/National Library of Medicine/National
Institutes of Health, the National Eye Institute/National Institutes of
Health, Department of Health and Human Services, Bethesda Maryland
(Contract HHS-N-260-2005-00007-C; ADB contract NO1-EY-5-0007).
Funds were generously contributed to these contracts by the following
National Institutes of Health: Office of Dietary Supplements, National
Center for Complementary and Alternative Medicine; National Institute on
Aging; National Heart, Lung, and Blood Institute; and National Institute of
Neurological Disorders and Stroke. The sponsor and funding organization
participated in the design and conduct of the study; data collection, management,
analysis and interpretation; and the preparation, review and
approval of the manuscript.
\vspace{1em}





\pagebreak
\parindent=2em

\section*{Abstract}
\vspace{-1.5em}
\hspace*{2em}\textbf{Purpose}: In assessing the severity of age-related macular degeneration (AMD), the Age-Related Eye Disease Study (AREDS) Simplified Severity Scale predicts the risk of progression to late AMD. However, its manual use requires the time-consuming participation of expert practitioners. Although several automated deep learning systems have been developed for classifying color fundus photographs (CFP) of individual eyes by AREDS severity score, none to date has used a patient-based scoring system that uses images from both eyes to assign a severity score.

\textbf{Design}: DeepSeeNet, a deep learning model, was developed to classify patients automatically by the AREDS Simplified Severity Scale (score 0-5) using bilateral CFP.

\textbf{Participants}: DeepSeeNet was trained on 58,402 and tested on 900 images from the longitudinal follow-up
of 4,549 participants from AREDS. Gold standard labels were obtained using reading center grades.

\textbf{Methods}: DeepSeeNet simulates the human grading process by first detecting individual AMD risk factors
(drusen size, pigmentary abnormalities) for each eye and then calculating a patient-based AMD severity score
using the AREDS Simplified Severity Scale. 

\textbf{Main Outcome Measures}: Overall accuracy, specificity, sensitivity, Cohen's kappa, and area under the
curve (AUC). The performance of DeepSeeNet was compared with that of retinal specialists.

\textbf{Results}: DeepSeeNet performed better on patient-based classification (accuracy = 0.671; kappa = 0.558)
than retinal specialists (accuracy = 0.599; kappa = 0.467) with high AUC in the detection of large drusen (0.94),
pigmentary abnormalities (0.93), and late AMD (0.97). DeepSeeNet also outperformed retinal specialists in the
detection of large drusen (accuracy 0.742 vs. 0.696; kappa 0.601 vs. 0.517) and pigmentary abnormalities (accuracy
0.890 vs. 0.813; kappa 0.723 vs. 0.535) but showed lower performance in the detection of late AMD
(accuracy 0.967 vs. 0.973; kappa 0.663 vs. 0.754). 

\textbf{Conclusions}: By simulating the human grading process, DeepSeeNet demonstrated high accuracy with
increased transparency in the automated assignment of individual patients to AMD risk categories based on the
AREDS Simplified Severity Scale. These results highlight the potential of deep learning to assist and enhance
clinical decision-making in patients with AMD, such as early AMD detection and risk prediction for developing late
AMD. DeepSeeNet is publicly available on \url{https://github.com/ncbi-nlp/DeepSeeNet}.

\pagebreak

Age-related macular degeneration (AMD) is responsible for
approximately 9\% of global blindness and is the leading
cause of visual loss in developed countries.\cite{quartilho2016leading,congdon2004causes} 
The number
of people with AMD worldwide is projected to be 196
million in 2020, increasing substantially to 288 million in
2040.\cite{wong2014global}
The prevalence of AMD increases exponentially
with age: late AMD in white populations has been
estimated by meta-analysis at 6\% at 80 years and 20\% at
90 years.\cite{rudnicka2012age}
Over time, increased disease prevalence through
changing population demographics may place great
burdens on eye services, especially where retinal
specialists are not available in sufficient numbers to
perform individual examinations on all patients. It is
conceivable that deep learning or telemedicine approaches
might support future eye services; however, this might
only apply when evidence-based systems have undergone
extensive validation and demonstrated performance metrics
that are at least noninferior to those of clinical ophthalmologists
in routine practice.

Age-related macular degeneration arises from a complex
interplay among aging, genetics, and environmental risk
factors.\cite{fritsche2014age,ratnapriya2013age}
It is regarded as a progressive, stepwise disease
and is classified by clinical features (based on clinical
examination or color fundus photography) into early,
intermediate, and late stages.\cite{ferris2013clinical}
The hallmarks of intermediate
disease are the presence of large drusen or pigmentary
abnormalities at the macula. There are 2 forms of late
AMD: (1) neovascular AMD and (2) atrophic AMD, with
geographic atrophy (GA).

The Age-Related Eye Disease Study (AREDS), sponsored
by the National Eye Institute (National Institutes of
Health), was a randomized clinical trial to assess the effects
of oral supplementation with antioxidant vitamins and
minerals on the clinical course of AMD and age-related
cataract. Longitudinal analysis of this study cohort led to
the development of the patient-based AREDS Simplified
Severity Scale for AMD, based on color fundus photographs.\cite{group2005simplified}
This simplified scale provides convenient risk
factors for the development of advanced AMD that can be
determined by clinical examination or by less demanding
photographic procedures than used in the AREDS. The
scale combines risk factors from both eyes to generate an
overall score for the individual, based on the presence of
1 or more large drusen (diameter~$>125 \mu m$) or AMD
pigmentary abnormalities at the macula of each eye.\cite{group2005simplified}
The
Simplified Severity Scale is also clinically useful in that it
allows ophthalmologists to predict an individual's 5-year
risk of developing late AMD. This 5-step scale (from
score 0 to 4) estimates the 5-year risk of the development of
late AMD in at least 1 eye as 0.4\%, 3.1\%, 11.8\%, 25.9\%,
and 47.3\%, respectively.\cite{group2005simplified}

Automated image analysis tools have demonstrated
promising results in biology and medicine.\cite{ching2018opportunities,wang2017chestx,wang2018tienet,banerjee2017intelligent,esteva2017dermatologist,ehteshamibejnordi2017diagnostic,lehman2015diagnostic}
In particular,
deep learning, a subfield of machine learning, has recently
generated substantial interest in the field of ophthalmology.\cite{ching2018opportunities,grassmann2018deep,kermany2018identifying,burlina2016detection,burlina2017automated,lam2018retinal,lee2017deep}
Past studies have used deep learning systems
for the identification of various retinal diseases, including
diabetic retinopathy,~\cite{choi2017multi,gargeya2017automated,gulshan2016development,raju2017development,takahashi2017applying,ting2017development}
glaucoma,\cite{ting2017development,asaoka2016detecting,cerentini2017automatic,muhammad2017hybrid}
retinopathy of prematurity,\cite{brown2018automated}
and AMD.\cite{burlina2017automated,lee2017deep,ting2017development,matsuba2018accuracy,treder2018automated}
In general, deep
learning is the process of training algorithmic models with
labeled data (e.g., color fundus photographs categorized
manually as containing pigmentary abnormalities or not),
where these models can then be used to assign labels
automatically to new data. Deep learning differs from
traditional machine learning methods in that specific image
features do not need to be prespecified by experts in that
field. Instead, the image features are learned directly from
the images themselves.

Recently, several deep learning systems have been
developed for the classification of color fundus photographs
into AMD severity scales, at the level of the individual eye.
These severity scales have included both binary (e.g.,
referable vs. nonreferable AMD\cite{kermany2018identifying,burlina2017automated,lee2017deep,ting2017development,matsuba2018accuracy}) and multi-class
(e.g., the 9-step AREDS Severity Scale\cite{grassmann2018deep,burlina2018use} and a 4-class
AMD classification\cite{burlina2017comparing}) systems. However, to the best of
our knowledge, none to date has developed a patientbased
system that, similar to the AREDS Simplified
Severity Scale score, uses images from both eyes to obtain
one overall score for the individual. This is particularly
relevant because estimates of rates of progression to late
AMD are highly influenced by the status of fellow eyes,
because the behavior of the 2 eyes is highly correlated.\cite{group2005simplified}
Additionally, several recent studies have reported robust
performance in the automated classification of AMD from
OCT scans.\cite{lee2017deep,karri2017transfer,defauw2018clinically,srinivasan2014fully,farsiu2014quantitative}
Unlike these studies, DeepSeeNet is
based on data from color fundus photography, which
remains an important imaging modality for assessing the
ophthalmic disease and is essential in grading eyes using the
AREDS Simplified Severity Score.\cite{marmor2011fluorescein}
Similar to the study by
De Fauw et al,\cite{defauw2018clinically} DeepSeeNet contains 2 stages by design
for improved performance and increased transparency.
However, their 2-stage approach is different from ours
with respect to the actual approach details as well as issues
in data variability.

The primary aim of our study was to train and test a
deep learning model to identify patient-level AMD severity
using the AREDS Simplified Severity Scale from color
fundus images of both eyes. Images were obtained from
the AREDS dataset, one of the largest available datasets
containing approximately 60,000 retinal images. Different
from previous methods, our model mimics the human
grading process by first detecting individual risk factors
(drusen and pigmentary abnormalities) in each eye and
then combining values from both eyes to assign an AMD
score for the patient. Thus, our model closely matches the
clinical decision-making process, which allows an
ophthalmologist to inspect and visualize an interpretable
result, rather than being presented with an AMD score by
a ``black-box'' approach. This approach offers potential
insights into the decision-making process, in a fashion
more typical of clinical practice, and has the advantages of
transparency and explainability.

\section*{Methods and Materials}

The specific aims of the study were (1) to compare the performance
of 3 deep learning models generated by 3 different training strategies;
and (2) for the most accurate of these 3 models, to compare
its performance with that of retinal specialists (AREDS investigators
whose assessments had previously been recorded during
the AREDS).

The reference measure used as the ``gold standard'' for both
training purposes and the measurement of performance was the
grading previously assigned to each color fundus photograph by
human graders at the Reading Center for the AREDS, as described
next.

\subsection*{Assignment of the AREDS Simplified Severity Scale by Reading Center grading}

This study used the AREDS dataset.\cite{group2005simplified} Briefly, the AREDS was a
12-year multi-center, prospective cohort study of the clinical
course, prognosis, and risk factors of AMD and age-related
cataract. Institutional review board approvals were obtained
from all 11 clinical sites, and written informed consents were
obtained from all AREDS participants. Stereoscopic color fundus
photographs from both eyes (field 2, 30\textdegree~imaging field centered at
the fovea) were obtained at the study baseline, the 2-year followup
visit, and annually thereafter. Because of the inherent redundancy
in a pair of stereoscopic photographs, for each eye, only 1
of the pair of photographs was used in the current study. In
general, the left image of the pair was used unless missing from
the database, in which case the right image was used instead
($\sim 0.5\%$).

The gold standard annotation (image labeling) was performed
by expert human graders at the Reading Center (University of
Wisconsin). The workflow is described in detail in AREDS Report
number 6.\cite{group2001age} In brief, a senior grader (grader 1) performed
preliminary grading of the photograph for AMD severity using a
standardized protocol for a 4-category scale, and a junior grader
(grader 2) performed detailed grading of the photograph for multiple
specific AMD features. A computerized algorithm then
extracted the AMD severity levels from the detailed gradings (by
grader 2). In the case of any discrepancy regarding the AMD
severity level between the graders, a senior investigator would
adjudicate the final severity level. All photographs were graded
independently, that is, graders were masked to the photographs and
grades from previous visits. Senior graders had approximately 10
to 15 years of experience, and junior graders had up to 5 years of
experience.

In addition, a rigorous process of grading quality control was
performed at the Reading Center including the assessment for the
inter-grader and intra-grader agreement overall and according to
specific AMD features.\cite{group2001age} Analyses for potential ``temporal drift''
were conducted by having all graders re-grade in a masked
fashion the same group of images annually for the duration of the
study.

\begin{figure}[H]
    \centering
    \frame{\includegraphics[clip,trim=8cm 0 9cm 0,width=.6\textwidth]{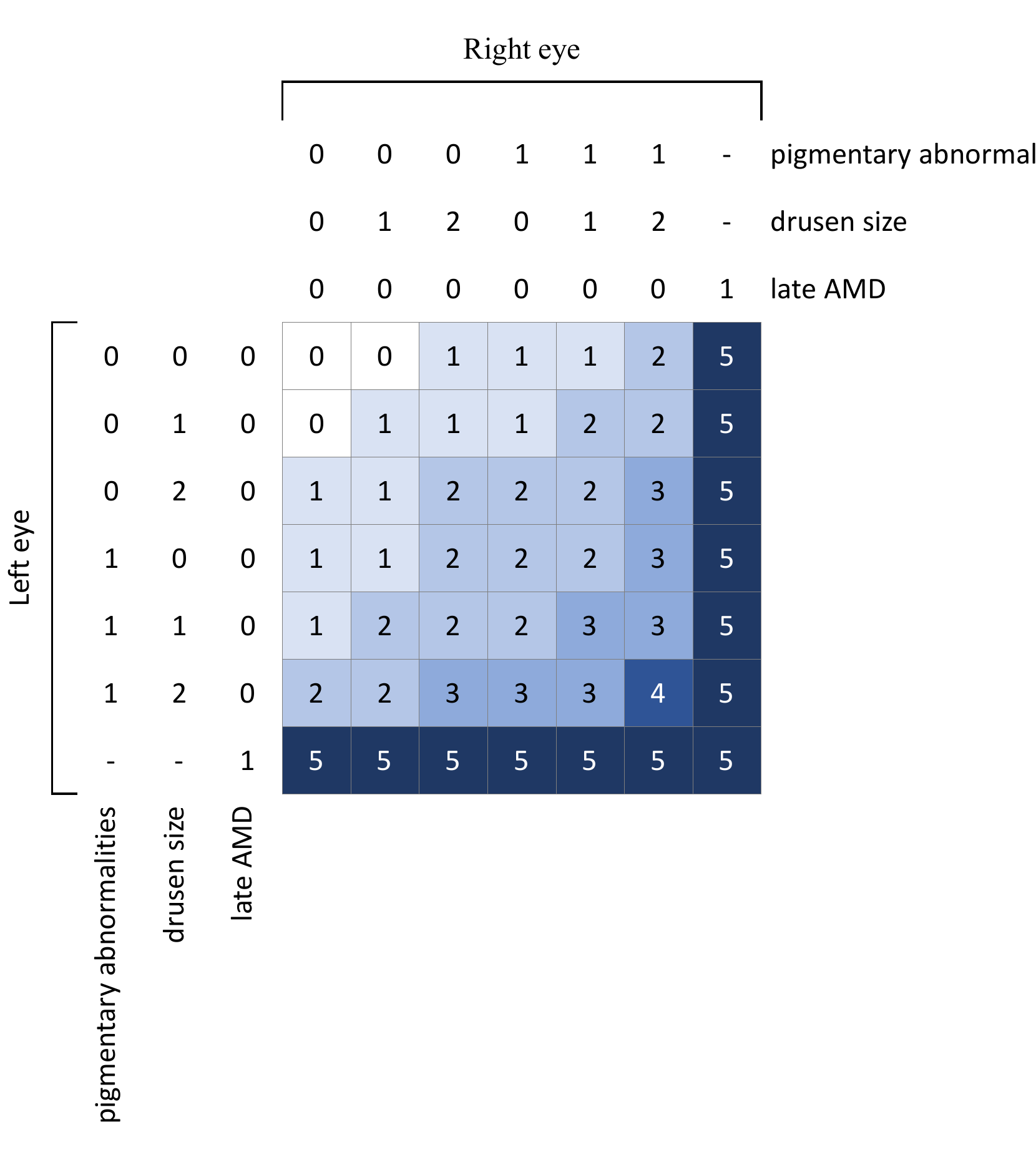}}
    \caption{Scoring schematic for participants with and without late agerelated
macular degeneration (AMD). Pigmentary abnormalities: 0 = no,
1 = yes; drusen size: 0 = small or none, 1 = medium, 2 = large; late AMD:
0 = no, 1 = yes.}
    \label{fig:scoring schematic}
\end{figure}

For each participant, at each time point, grades for both eyes
were used to calculate the AREDS Simplified Severity Scale score.
This scale ranges from 0 to 5, with a score of 0 to 4 assigned to
participants based on the drusen/pigment status in each eye, and a
score of 5 assigned to participants with late AMD (defined as
neovascular AMD or central GA) in either eye (Fig.~\ref{fig:scoring schematic}). This is a
modification of the original scoring method described by Ferris
et al.~\cite{group2005simplified} As described previously, these scores were used as gold
standard labels (i.e., reference), both for training purposes and to
assess the performance of the different models developed in this
study.

\subsection*{Image datasets used in the training and testing of the deep learning model}

The AREDS dataset is publicly accessible to researchers by request
at dbGAP (\url{https://www.ncbi.nlm.nih.gov/projects/gap/cgi-bin/study.cgi?study_id=phs000001.v3.p1}).\cite{group2005simplified}
A total of 59,302 color
fundus images from 4,549 participants were extracted from the
AREDS dataset. This dataset was divided into 2 subsets: (1) a
testing dataset, which consisted of bilateral images captured at
the study baseline from 450 participants (i.e., 1 image from each
eye); at the time of the study, in addition to undergoing normal
Reading Center grading, these images were also assessed
(separately and independently) by the retinal specialists, whose
responses were recorded; and (2) a training dataset, which
consisted of 58,402 images from the remaining 4,099 participants,
captured at multiple study visits (although not all participants
had follow-up visits through to 12 years). The images taken from
the group of 450 participants at visits other than the baseline visit
were not used in either dataset (Fig.~\ref{figs:training testing sets}). 
Table~\ref{tab:areds} summarizes the distribution of
participants by the AREDS Simplified Severity Scale at baseline.
Table~\ref{tab:risk factor} summarizes the distributions of scored AMD features
among the training and testing datasets.

\begin{table}[ht]
\centering
\begin{threeparttable}
\caption{Summary of Age-Related Eye Disease Study Participants
According to Age-Related Eye Disease Study Simplified Severity
Scale Scores at Study Baseline (by Reading Center Grading)}
\label{tab:areds}
\begin{tabular}{c@{\hspace{4em}}rrr@{\hspace{4em}}rr}
    \toprule
    AREDS Simplified & \multicolumn{5}{c}{No. of Participants (\% Total)}\\
    \cmidrule{2-6}
    Severity Scale Score & \multicolumn{2}{c}{Training} && \multicolumn{2}{c}{Testing}\\
    \midrule
    0 & 1,258 & (30.7) && 185 & (41.1)\\
    1 & 653 & (15.9) && 79 & (17.6)\\
    2 & 461 & (11.3) && 56 & (12.4)\\
    3 & 303 & (7.4) && 46 & (10.2)\\
    4 & 279 & (6.8) && 33 & (7.3)\\
    5 & 537 & (13.1) && 51 & (11.3)\\
    \midrule
    Total participants & 4,099 & (100.0) && 450 & (100.0)\\
    \bottomrule
\end{tabular}
\begin{tablenotes}
\item AREDS = Age-Related Eye Disease Study.
\end{tablenotes}
\end{threeparttable}
\end{table}
\begin{table}[ht]
\centering
\begin{threeparttable}
\caption{Number of Color Fundus Images in the Training and
Testing Sets Stratified by Risk Factors and Late Age-related
Macular Degeneration Categorization}
\label{tab:risk factor}
\centering
\begin{tabular}{l@{\hspace{4em}}rrr@{\hspace{4em}}rr}
    \toprule
    \multirow{2}{*}{Risk factors} & \multicolumn{5}{c}{Number of Fundus Images (\% Total)}\\
    \cmidrule{2-6}
     & \multicolumn{2}{c}{Training (all visits)} && \multicolumn{2}{c}{Testing (baseline)}\\
    \midrule
    Drusen & \\
    \hspace{2em} Small/none & 23,625  & (40.5) && 395 & (43.9)\\
    \hspace{2em} Medium & 16,020  & (27.4) && 206 & (22.9)\\
    \hspace{2em} Large & 18,757  & (32.1) && 299 & (33.2)\\
    \multicolumn{3}{l}{Pigmentary abnormalities} \\
    \hspace{2em} No & 36,712  & (62.9) && 631 & (70.1)\\
    \hspace{2em} Yes & 21,690  & (37.1) && 269 & (29.9)\\
    Late AMD & \\
    \hspace{2em} No & 50,800  & (87.0) && 849 & (94.3)\\
    \hspace{2em} Yes & 7,602  & (13.0) && 51 & (5.7)\\
    \midrule
    Total images & 58,402  & (100.0) && 900 & (100.0)\\
    \bottomrule
\end{tabular}
\begin{tablenotes}
\item AMD = age-related macular degeneration.
\end{tablenotes}
\end{threeparttable}
\end{table}

\subsection*{Composition of the DeepSeeNet deep learning model}

DeepSeeNet was designed as a deep learning model that could be
used to assign patient-based AREDS Simplified Severity Scale
scores in an automated manner using bilateral color fundus photographs (Fig.~\ref{figs:workflow}). DeepSeeNet
simulates the grading process of ophthalmologists by first
detecting the presence or absence of AMD risk-associated features
for each eye (large drusen and AMD pigmentary abnormalities)
and then using these bilateral data to compute a patient-based
score (0-5) using the algorithm described earlier.

DeepSeeNet consists of 3 constituent parts that contribute to its
output: (a) a sub-network, Drusen-Net (D-Net), which detects
drusen in 3 size categories (small/none, medium, and large); (b) a
sub-network, Pigment-Net (P-Net), which detects the presence or
absence of pigmentary abnormalities (hypopigmentation or hyperpigmentation);
and (c) a sub-network, Late AMD-Net (LA-Net),
which detects the presence or absence of late AMD (neovascular
AMD or central GA).

D-Net, P-Net, and LA-Net were designed as deep convolutional
neural networks (CNNs),\cite{lecun1989generalization} each with an Inception-v3 architecture,
\cite{szegedy2016rethinking} which is a state-of-the-art CNN model for image classification.
In total, there are 317 layers in the Inception-v3 model,
comprising a total of $>21$ million weights (learnable parameters)
that were subject to training.

Before training, we followed the lead of Burlina et al\cite{burlina2016detection,burlina2017comparing} to
preprocess our image data as follows: the AREDS fundus photographs
were cropped to generate a square image field encompassing
the macula, followed by scaling the image to a resolution
of $224 \times 224$ pixels (Fig.~\ref{figs:preprocessing}). We
trained our model in Keras with TensorFlow as the backend.\cite{abadi2016tensorflow,chollet2015}
During the training process, we updated the model parameters
using the Adam optimizer (learning rate of 0.0001) for every
minibatch of 32 images.\cite{kingma2015adam} This reduces the variance of the
parameter update, which leads to a more stable convergence. The
training was stopped after 5 epochs (passes of the entire training
set) once the accuracy values no longer increased or started to
decrease. All experiments were conducted on a server with 32
Intel Xeon CPUs, using a NVIDIA GeForce GTX 1080 Ti 11Gb
GPU for training and testing, with 512 Gb available in RAM
memory.

\subsection*{Performance comparison between DeepSeeNet and retinal specialists }

We compared the performance of the deep learning model with that
of retinal specialists, using the Reading Center grades as the gold
standard, in both cases. For the performance of the retinal specialists,
we used the AREDS Simplified Severity Scale scores that
had previously been recorded from the retinal specialists who
originally served as the AREDS investigators. These scores were
recorded at the AREDS baseline study visits, when the retinal
specialists (n = 88) had independently assessed 450 AREDS
participants as part of a qualification survey used to determine
initial AMD severity for each eye. The clinical assessment
involved the determination of the following features: drusen size
(within 2 disc diameter of the macula center), presence of
pigmentary abnormalities consistent with AMD (within 1 disc
diameter), AMD subretinal neovascularization, previous laser
photocoagulation for AMD subretinal neovascularization, central
GA, retinal pigment epithelial detachment, and disciform scar.
These clinical assessments were used to derive the same patient-based
Simplified Severity Scale as defined in Fig.~\ref{fig:scoring schematic}.

Overall accuracy, specificity, sensitivity, Cohen's kappa,\cite{cohen1968multiple,cohen2016coefficient}
and receiver operating characteristic curve analysis were used to
evaluate the performance of DeepSeeNet and retinal specialists
(with reference to the Reading Center grades as the gold standard).
Kappa values $<0$ indicate no agreement, 0 to 0.20 indicate slight
agreement, 0.21 to 0.40 indicate fair agreement, 0.41 to 0.60
indicate moderate agreement, 0.61 to 0.80 indicate substantial
agreement, and 0.81 to 1 indicate almost perfect agreement.\cite{landis1977measurement} We
also followed the work of Poplin et al\cite{poplin2018prediction} to assess the statistical
significance of the results. For the test dataset, we sampled 450
patients with replacement and evaluated the model on this
sample. By repeating this sampling and evaluation 2,000 times,
we obtained a distribution of the performance metric (e.g.,
kappa) and reported 95\% confidence intervals.

\section*{Results}

\subsection*{Predicting AREDS simplified severity scale}

DeepSeeNet predicted AREDS Simplified Severity Scale scores for
each participant in the testing dataset (n = 450). The performance
of the deep learning models was measured against the Reading
Center grades previously assigned to these 450 participants (as the
reference or gold standard).

We investigated 3 strategies for training and optimizing
DeepSeeNet (details located under ``Training Strategies'' in the
in Appendix 1, Fig.~\ref{figs:Different deep learning algorithmic}, and Table~\ref{tabs:Performance of three different deep learning models})
and found the fine-tuning strategy (all layers in a pretrained
Inception-v3 model were fine-tuned using the AREDS dataset)
achieved the best results, with accuracy = 0.671 and kappa =
0.558. As a result, we will discuss only Fine-tuned DeepSeeNet
hereafter.

The performance of Fine-tuned DeepSeeNet was then
compared with that of the retinal specialists (Table~\ref{tab:performance}). The
performance of DeepSeeNet (accuracy = 0.671; kappa = 0.558)
was superior to that of the retinal specialists (accuracy = 0.599;
kappa = 0.467).
\begin{table}[ht]
\centering
\begin{threeparttable}
\caption{Performance of Fine-tuned DeepSeeNet Compared with
Retinal Specialists on Classifying Age-Related Eye Disease Study
Simplified Severity Scale Scores from Color Fundus Photographs}
\label{tab:performance}
\begin{tabular}{lrr}
    \toprule
     & Fine-tuned DeepSeeNet & Retinal specialist\\
     \cmidrule(r){2-2} \cmidrule(l){3-3}
     & (95\% CI) & (95\% CI)\\
    \midrule
    Overall accuracy  & 0.671 (0.670-0.672) & 0.599 (0.598-0.600)\\
    Sensitivity       & 0.590 (0.589-0.591) & 0.512 (0.511-0.513)\\
    Specificity       & 0.930 (0.930-0.930) & 0.916 (0.916-0.916)\\
    Kappa             & 0.558 (0.557-0.560) & 0.467 (0.466-0.468)\\
    \bottomrule
\end{tabular}
\begin{tablenotes}
\item CI = confidence interval.
\end{tablenotes}
\end{threeparttable}
\end{table}

\begin{figure}[H]
    \centering
    \frame{\includegraphics[clip,trim=0 7cm 0 7cm,width=\textwidth]{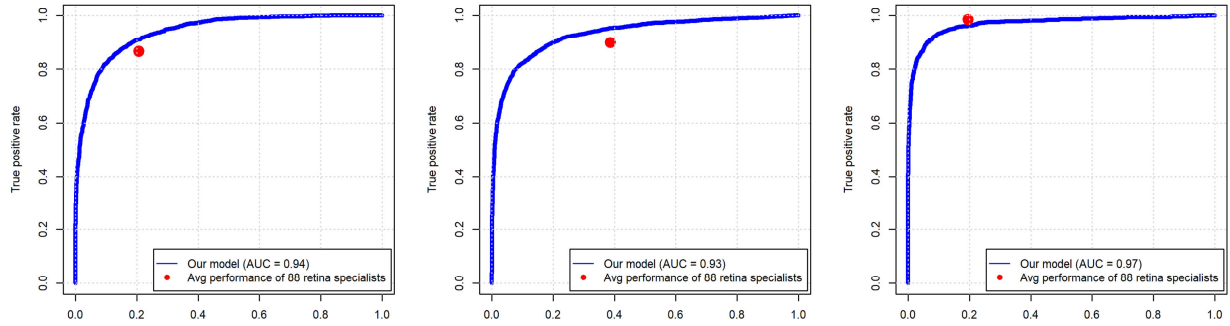}}
    \caption{Receiver operating characteristic curves for large drusen, pigmentary abnormalities, and late age-related macular degeneration (AMD) classification.
Retinal specialists' performance levels are represented as a single red point. AUC = area under the curve.}
    \label{fig:roc}
\end{figure}

In addition, the performance of the individual sub-networks
used in Fine-tuned DeepSeeNet (D-Net, P-Net, and LA-Net) was
compared with that of retinal specialists (Table~\ref{tab:Performance of risk factor prediction}). Fig.~\ref{fig:roc}
displays receiver operator characteristic curves for the individual
sub-networks, with the average performance of the retinal specialists
shown as single red points. The performance of D-Net and
P-net was superior to the performance of the retinal specialists in
assessing large drusen and pigmentary abnormalities, respectively.
The accuracy of LA-Net was similar to that of the retinal specialists
in assessing the presence of late AMD, but its kappa was lower.
\begin{landscape}
\begin{table}[p]
\centering
\begin{threeparttable}
\caption{Performance of Risk Factor Prediction (Retinal Specialists vs. Individual Sub-Network Models)}
\label{tab:Performance of risk factor prediction}
\footnotesize
\begin{tabular}{lcccccc}
    \toprule
     & \multicolumn{2}{c}{Drusen} & \multicolumn{2}{c}{Pigmentary Changes} & \multicolumn{2}{c}{Late AMD}\\
    \cmidrule(r){2-3}\cmidrule(r){4-5}\cmidrule(r){6-7}
     & Retinal specialist & D-Net & Retinal specialist & P-Net & Retinal specialist & LA-Net\\
    \midrule
    Overall accuracy (95\% CI) & 0.696 (0.695-0.697) & 0.742 (0.741-0.742) & 0.813 (0.813-0.814) & 0.890 (0.889-0.890) & 0.973 (0.973-0.973) & 0.967 (0.967-0.967)\\
    Sensitivity (95\% CI) & 0.635 (0.634-0.636) & 0.718 (0.717-0.719) & 0.615 (0.613-0.616) & 0.732 (0.731-0.733) & 0.801 (0.798-0.805) & 0.627 (0.626-0.632)\\
    Specificity (95\% CI) & 0.842 (0.842-0.843) & 0.871 (0.871-0.872) & 0.898 (0.898-0.899) & 0.957 (0.957-0.957) & 0.983 (0.983-0.984) & 0.987 (0.987-0.987)\\
    Kappa (95\% CI) & 0.517 (0.516-0.518) & 0.601 (0.600-0.602) & 0.535 (0.533-0.536) & 0.723 (0.722-0.724) & 0.754 (0.751-0.757) & 0.663 (0.660-0.665)\\
    \bottomrule
\end{tabular}
\begin{tablenotes}
\item AMD = age-related macular degeneration; CI = confidence interval; D-Net = Drusen-Net, which classifies drusen into 3 size categories (small/none, medium, and large); LA-Net = Late AMD-Net, which
detects the presence or absence of late AMD (neovascular AMD or central geographic atrophy [GA]); P-Net = Pigment-Net, which detects the presence or absence of any pigmentary abnormality consistent
with AMD (hypopigmentation or hyperpigmentation).
\end{tablenotes}
\end{threeparttable}
\end{table}
\end{landscape}

\begin{figure}[H]
    \centering
    \frame{\includegraphics[clip,trim=2cm 2cm 2cm 2cm,width=.8\textwidth]{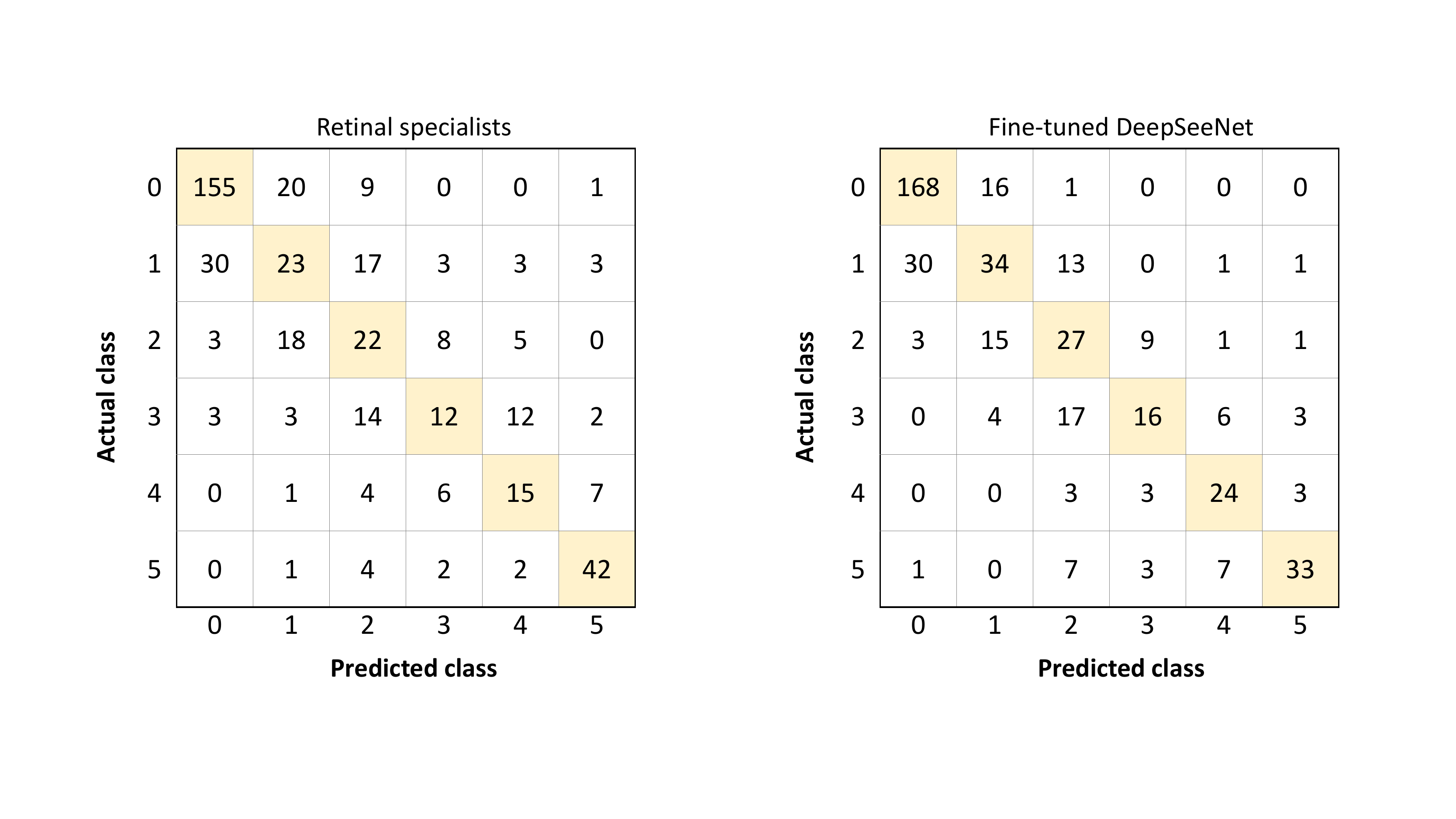}}
    \caption{Confusion matrices comparing retinal specialists' performance with that of DeepSeeNet based on the test set values. The rows and columns of
each matrix are the Scale scores (0-5).}
    \label{fig:confusion}
\end{figure}

Fig.~\ref{fig:confusion} shows confusion matrices comparing the performance
of Fine-tuned DeepSeeNet and the retinal specialists in grading
AMD severity (with accuracy comparisons detailed in Table~\ref{tabs:Accuracy comparing retinal}). These matrices depict the true
versus the predicted AREDS Simplified Severity Scale scores of
the 450 participants at baseline. The numbers of predictions are
summarized with count values broken down by each class,
indicating the accuracy and errors made by DeepSeeNet or the
retinal specialists. Fig.~\ref{fig:confusion} shows that DeepSeeNet correctly
classified scores 0 to 4 more often than the retinal specialists,
whereas the retinal specialists correctly classified late AMD more
often than DeepSeeNet.

Last, the performance of Fine-tuned DeepSeeNet on all images
in the test set (accuracy = 0.662; kappa = 0.555) was compared
with that on the images at study baseline only (Table~\ref{tabs:Performance of DeepSeeNet on assessing} and Table~\ref{tabs:Performance of three different deep learning models}). We observed that the
accuracy on the study baseline was slightly better, though the
kappas values were similar. Although the distribution of AMD
severity (for the testing cases) was slightly less severe for the
study baseline images, we do not consider this to have
introduced bias because the test cases were the same for the
model as for the retinal specialists.

\subsection*{Interpretation}

Although Fine-tuned DeepSeeNet demonstrated relatively robust
performance on classifying color fundus photographs according to
AMD severity, the mechanics of this and other deep learning
models are sometimes considered cryptic or lacking in transparency.
Indeed, for this reason, deep learning models are often
referred to as ``black-box'' entities. To improve transparency, in
addition to creating models composed of sub-networks with overt
purposes, we applied 2 additional techniques to aid interpretation
of the results.

\subsection*{T-Distributed Stochastic Neighbor Embedding (t-SNE) Method}

\begin{figure}[H]
    \centering
    \frame{\includegraphics[trim=0 3cm 0 3cm,clip,width=\textwidth]{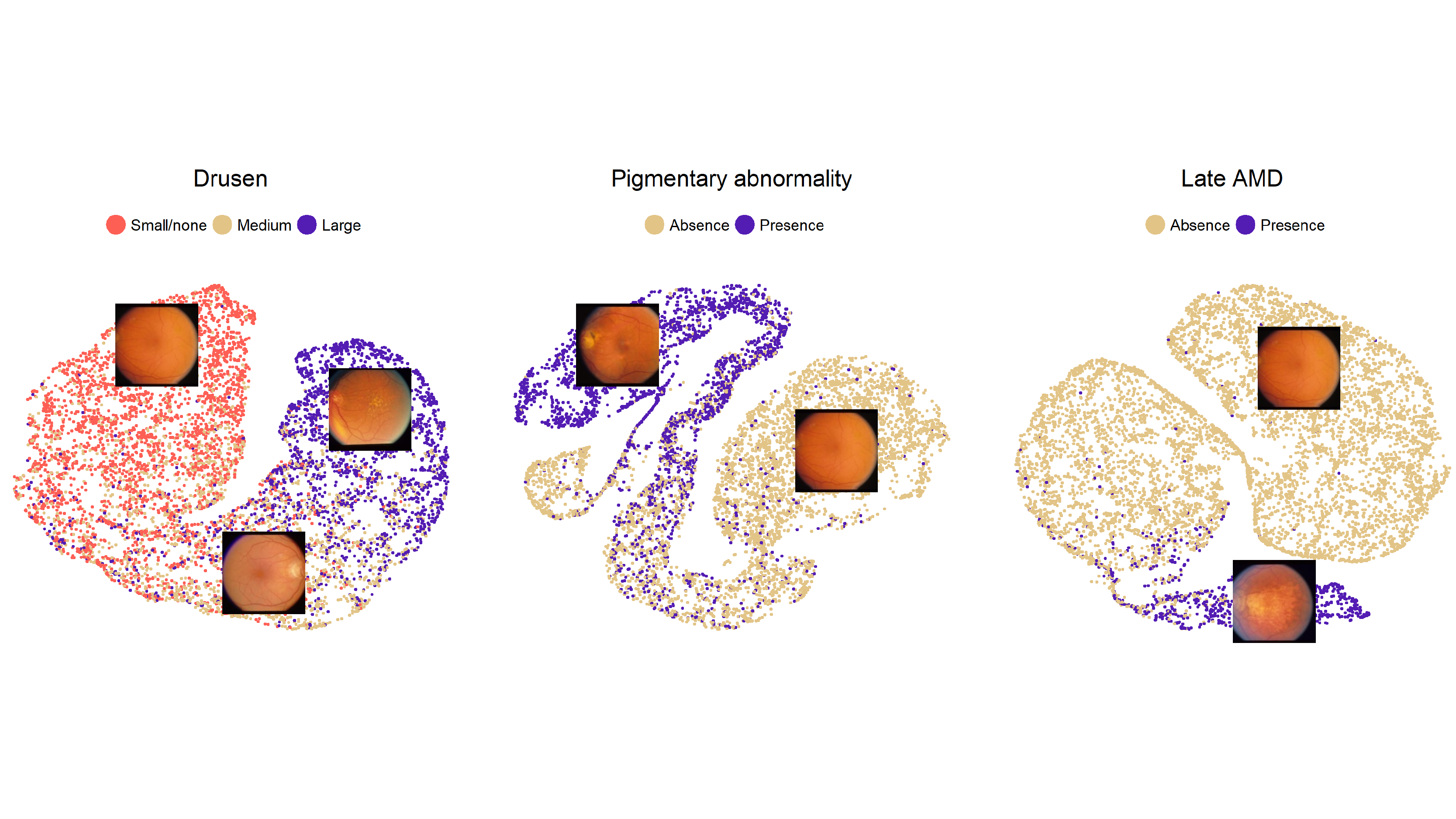}}
    \caption{The t-distributed Stochastic Neighbor Embedding (t-SNE) visualization of the last hidden layer representation for each sub-network of Deep-
SeeNet. Each point represents a fundus image. Different colors represent the different classes of the respective risk factor or late age-related macular
degeneration (AMD).}
    \label{fig:t-sne}
\end{figure}

In this study, the internal features learned by Fine-tuned DeepSeeNet were studied using t-distributed Stochastic Neighbor
Embedding (see Glossary in Table 5), which is well suited for the
visualization of high-dimensional datasets.\cite{simonyan2013deep} We first obtained the
128-dimensional vector of DeepSeeNet's last dense layer and
applied the t-distributed Stochastic Neighbor Embedding technique
to reduce the vector into 2 dimensions for visualization (Fig.~\ref{fig:t-sne}).
Fig.~\ref{fig:t-sne} demonstrates that, for drusen, small/none drusen and
large drusen were split across the medium drusen point cloud.
The figure contains some points that are clustered with the
wrong class, many of which are medium drusen and difficult to
identify. For pigmentary abnormality and late AMD, presence
and absence classes were separated clearly.

\subsection*{Saliency Method}

\begin{figure}[H]
    \centering
    \frame{\includegraphics[clip,trim=1cm 0 2.5cm 2cm,width=\textwidth]{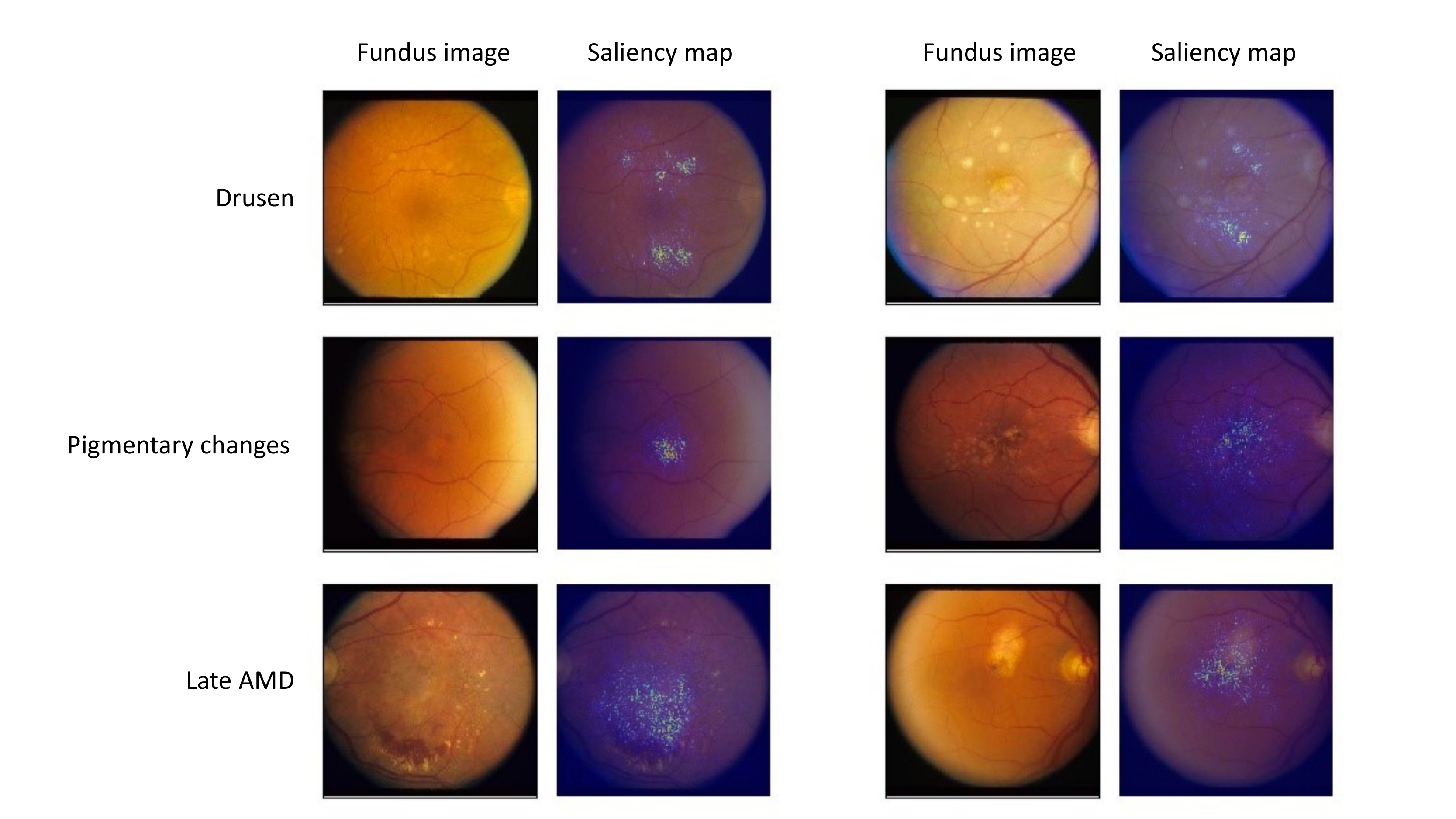}}
    \caption{Image-specific class saliency maps applied to 6 color fundus photographs: 2 eyes with large drusen, 2 eyes with pigmentary abnormalities, and 2
eyes with late age-related macular degeneration (AMD). In the saliency maps, the areas marked with bright signal correspond to the pixels that contributed
most to the model's classification of that class. The corresponding color fundus photograph is shown next to the saliency map for reference.}
    \label{fig:saliency}
\end{figure}

The second method used to aid interpretation of the results toward
model transparency was the saliency method. To visualize
important areas in the color fundus images (i.e., those areas that
contributed most toward classification), we applied image-specific
class saliency maps to assess manually whether DeepSeeNet was
concentrating on image areas that human experts would consider
the most appropriate to predict AMD severity.\cite{simonyan2013deep} The saliency map
is widely used to represent the visually dominant location in a
given image, corresponding to the category of interest, by
back-projecting the relevant features through the CNN. It helps
highlight areas used by the deep learning algorithm for prediction
and can provide insight into misclassified images. For example, as
seen in the ``drusen'' category of Fig.~\ref{fig:saliency}, the areas highlighted in
the saliency maps are indeed areas with drusen that are visually
apparent in the color fundus images. Likewise, in the
``pigmentary changes'' and ``late AMD'' categories in Fig.~\ref{fig:saliency},
the areas highlighted in the saliency maps are visually confirmed
to correspond with the relevant features in the corresponding
color fundus images. However, although saliency maps aid
interpretation by highlighting the dominant areas, they are
limited in that they do not completely explain how the algorithm
came to its final decision.

\section*{Discussion}

The accuracy of Fine-Tuned DeepSeeNet was superior to
that of human retinal specialists (accuracy 67\% vs. 60\%),
together with moderate agreement with the ground truth as
indicated by the kappa score. If deep learning approaches
were to support eye services in the future, comparisons of
this kind (with demonstration of noninferiority to human
clinicians) would be very important, together with extensive
validation across multiple and diverse image datasets. Of
note, although the overall accuracy of Fine-tuned
DeepSeeNet was superior, subgroup analysis showed that
Fine-tuned DeepSeeNet classified participants with Severity
Scale scores 0 to 4 correctly more often than the retinal
specialists, whereas the retinal specialists classified late
AMD correctly more often than Fine-Tuned DeepSeeNet
(Fig.~\ref{fig:confusion}). However, one important potential reason for the
latter difference is that the number of images of late AMD
that were available for model training was relatively low
at 13.0\% of the total training set (537 participants). We
postulate that further training of Fine-tuned DeepSeeNet
with larger numbers of late AMD images may improve its
performance in this area.

\subsection*{Error analysis on misclassified images in the AREDS testing dataset}

We considered that useful lessons might be learned by
careful examination of those instances where Fine-tuned
DeepSeeNet made errors in the Severity Scale classification,
particularly (as described earlier) in the case of late
AMD (where its accuracy was lower than that of the retinal
specialists). The matrices shown in Fig.~\ref{fig:confusion} demonstrate
that, for actual Severity Scale scores of 0 to 4, in the
large majority of cases, the score predicted by Fine-tuned
DeepSeeNet was incorrect by 1 scale step only.
We also examined those cases in which Fine-tuned
DeepSeeNet incorrectly classified a participant as having
late AMD (score 5) and found that, in 50\% of these cases,
noncentral GA was present in at least 1 eye. For the purposes
of this study, noncentral GA was not defined as late
AMD, although recent studies have expanded the definition
of late AMD to include noncentral GA.\cite{ferris2013clinical} The
misclassification of these images by our deep learning
model suggests an inherent similarity between these
groups of images.

Image quality also affected the accuracy of the deep
learning model. Of the participants classified incorrectly by
Fine-tuned DeepSeeNet as having late AMD, 25.0\% had
digital artifacts obscuring the fovea. In addition, image
brightness affected the model accuracy. Participants with a
pale retina or digital artifacts were more likely to be
misclassified as having GA. In the future, we aim to
address these problems by identifying color fundus photographs
with inferior quality, either for exclusion or for
additional processing.

\subsection*{Strengths, limitations, and future work}

One current limitation of DeepSeeNet (at least in its
present iteration) arises from the imbalance of cases
that were available in the AREDS dataset used for its
training, particularly the relatively low proportion of
participants with late AMD. As described previously,
this is likely to have contributed to the relatively lower
accuracy of DeepSeeNet in the classification of late
AMD, that is, through the performance of LA-Net in
the overall model. However, this limitation may
potentially be addressed by further training using image
datasets with a higher proportion of late AMD
cases.

A limitation of this dataset includes the sole use of
color fundus photographs because these were the only
images obtained in a study that began in 1992. Other
imaging techniques such as OCT and fundus autofluorescence
images were not yet feasible or universally
available. Future studies would benefit from inclusion of
additional methods of imaging. Multimodal imaging would
be desirable.

Another potential limitation lies in the reliance of
DeepSeeNet on higher levels of image quality for accurate
classification. Unlike in other studies,\cite{grassmann2018deep,burlina2017automated} we did not
perform extensive preprocessing of images, such as the
detection of the outer boundaries of the retina or normalization
of the color balance and local illumination. It is
possible that the use of these techniques might have
improved the accuracy of the model. However, we deliberately
avoided extensive preprocessing to make our model
as generalizable as possible.

We recommend further testing of our deep learning
model using other datasets of color fundus images. In
addition, it would be interesting for future studies to
compare the accuracy of the model with that of different
groups of ophthalmologists (e.g., retinal specialists, general
ophthalmologists, and trainee ophthalmologists).
Indeed, a recent study on grader variability for diabetic
retinopathy severity using color fundus photographs
suggested that retinal specialists have a higher accuracy
than that of general ophthalmologists.\cite{krause2018grader} In this study, we
set the bar as high as possible for the deep learning
model, because we considered that the retinal specialists
might have accuracy as close as possible to that of the
Reading Center gradings.

In conclusion, this study shows that DeepSeeNet performed
patient-based AMD severity classification with a
level of accuracy higher than a group of human retinal
specialists. If these results are tested and validated by
further reports of superiority across multiple datasets
(ideally from different countries), it is possible that the
integration of deep learning models into clinical practice
might become increasingly acceptable to patients and
ophthalmologists. In the future, deep learning models
might support eye services by reducing the time and human
expertise needed to classify retinal images and might lend
themselves well (through telemedicine approaches) to
improving care in geographical areas where current services
are absent or limited. Although deep learning models
are often considered ``black-box'' entities (because of
difficulties in understanding how algorithms make their
predictions), we aimed to improve the transparency of
DeepSeeNet by constructing it from sub-networks with
clear purposes (e.g., drusen detection) and analyzing its
outputs with saliency maps. These efforts to demystify
deep learning models may help improve levels of acceptability
to patients and adoption by ophthalmologists. We
have also analyzed the performance of several distinct
training strategies; lessons from these approaches may
have applicability to the development of deep learning
models for other retinal diseases, such as diabetic retinopathy,
and even for image-based deep learning systems
outside of ophthalmology.

Our new model uses deep learning in combination with a
clinically useful, patient-based, AMD classification system that
combines risk factors from both eyes to obtain a score for the
patient. The deep learning model and data partition are
publicly available (\url{https://github.com/ncbi-nlp/DeepSeeNet}).
By making these available, we aim to maximize the transparency
and reproducibility of this study, and to provide a
benchmark method for the further refinement and development
of methodologies. In addition, this deep learning model,
trained on one of the largest publicly available color fundus
photograph repositories, may allow for future deep learning
studies of other retinal diseases in which only smaller datasets
are currently available.

In the future, we aim to improve the model by incorporating
other information such as demographic, medical, and
genetic data, potentially together with imaging data from
other modalities. We also plan to evaluate our model on a
new dataset from the second AREDS (AREDS2) sponsored
by the National Eye Institute. In addition, we hope to
investigate the combination of OCT-based and color fundus
photographs-based deep learning models once each has
been more highly validated individually. Taken together, we
expect this study will contribute to the advancement and
understanding of retinal disease and may ultimately enhance
clinical decision-making.

\section*{Glossary}
\singlespacing
\begin{longtable}{p{.2\textwidth}p{.75\textwidth}}
\hline
Term & Description\\\hline
Adam optimizer & Adam is an optimization algorithm to update network weights. Different from classic optimization that maintains a single learning rate for all weight updates, with the learning rate not changing during training, it computes adaptive learning rates for different parameters during the training.\cite{kingma2015adam}.\\\hline
Back-propagation & A method used in artificial neural networks to calculate a gradient that is needed in the calculation of the weights to be used in the network~\cite{goodfellow2016deep}.\\\hline
Convolutional neural network & A class of artificial neural network algorithms utilized in deep learning largely for image classification. \\\hline
Deep learning & A subfield of machine learning in which explicit features are determined from the training data and do not require pre-specification by human domain experts.\\\hline
Epoch & A single pass through the entire training set. \\\hline
Fine-tune & A process to take a neural network model that has already been trained for a given task, and make it perform a second task.\\\hline
Fully-connected layer & A linear operation in which every output neuron has connections to all activations in the previous layer.\\\hline
Hidden layer & The middle layer of a neural network.\\\hline
ImageNet & An image database comprised of $>14$ million natural images and their corresponding labels. Due to the large number of labeled images, this dataset is often employed in deep learning techniques to pre-train models. In a process known as ``transfer learning'', the first layers are trained with ImageNet to extract more primitive features from the images (e.g., edge detection).\\\hline
Inception-v3 & A convolutional neural network with the inception architecture for computer vision~\cite{szegedy2016rethinking}.\\\hline
Layer & A container that usually receives weighted input, transforms it with a set of mostly non-linear functions, and then passes these values as output to the next layer. \\\hline
Leaning rate & A hyper-parameter that controls how much the weights of deep neural network are adjusted with respect the loss gradient.\\\hline
Multiclass classification & A classification task with more than two classes.\\\hline
Multilayer perceptron & A class of feedforward artificial neural network that consists of at least one hidden layer.\\\hline
Over-fitting & The production of an analysis that corresponds too closely or exactly to a particular set of data and may therefore fail to fit additional data or predict future observations reliably.\\\hline
Saliency map & The saliency map is computed for an input image and a given output class. It tells us which pixels in the image contribute most to the model's classification of that class. Specifically, we first computed the gradient of a given label with respect to the input image. The gradient, by definition, points in the direction of the greatest rate of class changes with respect to a small change in the input images. That small region of changes in the input image, thereby, contributes most and is highlighted in the saliency map. \\\hline
Stochastic gradient descent  & An iterative method for optimizing the objective function in machine learning.\\\hline
Test set & A mutually exclusive set of images not utilized in the training set. These images are used for testing deep learning models to evaluate their performance levels.\\\hline
Training & A data-driven approach requiring tens of thousands of labeled images in the training set.\\\hline
Training set & The set of images used for training a deep learning model. The network then predicts the category of each image and compares it with known ``ground truth'' labels. The parameters in the network are then optimized to improve the model's predictive ability, in a process known as ``back-propagation''.\\\hline
Transfer learning & The process of training a deep learning model on a large set of data, such that the model's weights are optimized as learned features. These weights are then ``transferred'' to a new neural network to allow for more efficient training of the model on a new training set (often smaller in size).\\\hline
t-distributed Stochastic Neighbor Embedding & A technique used to visualize and explore complex datasets (particularly those with high-dimensional features) in a low-dimensional space. In our case, we used it to create a 2-dimensional map by assigning a location to each datapoint (each retinal image). The locations are decided by probability distributions, such that datapoints that are similar across high-dimensional features end up close to each other, and datapoints that are dissimilar end up far apart. As a result, t-SNE plots often seem to display clusters (e.g., the cluster for large drusen, in this case), where the datapoints in the cluster all have relatively similar features. Therefore, it can be used to help the classification process and in the visual inspection and exploration of results from deep learning experiments.\cite{maaten2008visualizing}. \\\hline
Weights & Learnable parameters of the deep learning model.\\
\hline
\end{longtable}

\bibliographystyle{vancouver}
\bibliography{references}

\pagebreak

\doublespacing

\section*{Appendix 1}
\setcounter{figure}{0}
\setcounter{table}{0}
\renewcommand{\thefigure}{S\arabic{figure}}
\renewcommand{\thetable}{S\arabic{table}}

\subsection*{Training strategies}

\begin{figure}[!ht]
    \centering
    \frame{\includegraphics[clip,trim=4cm 4cm 4cm 0cm,width=\textwidth]{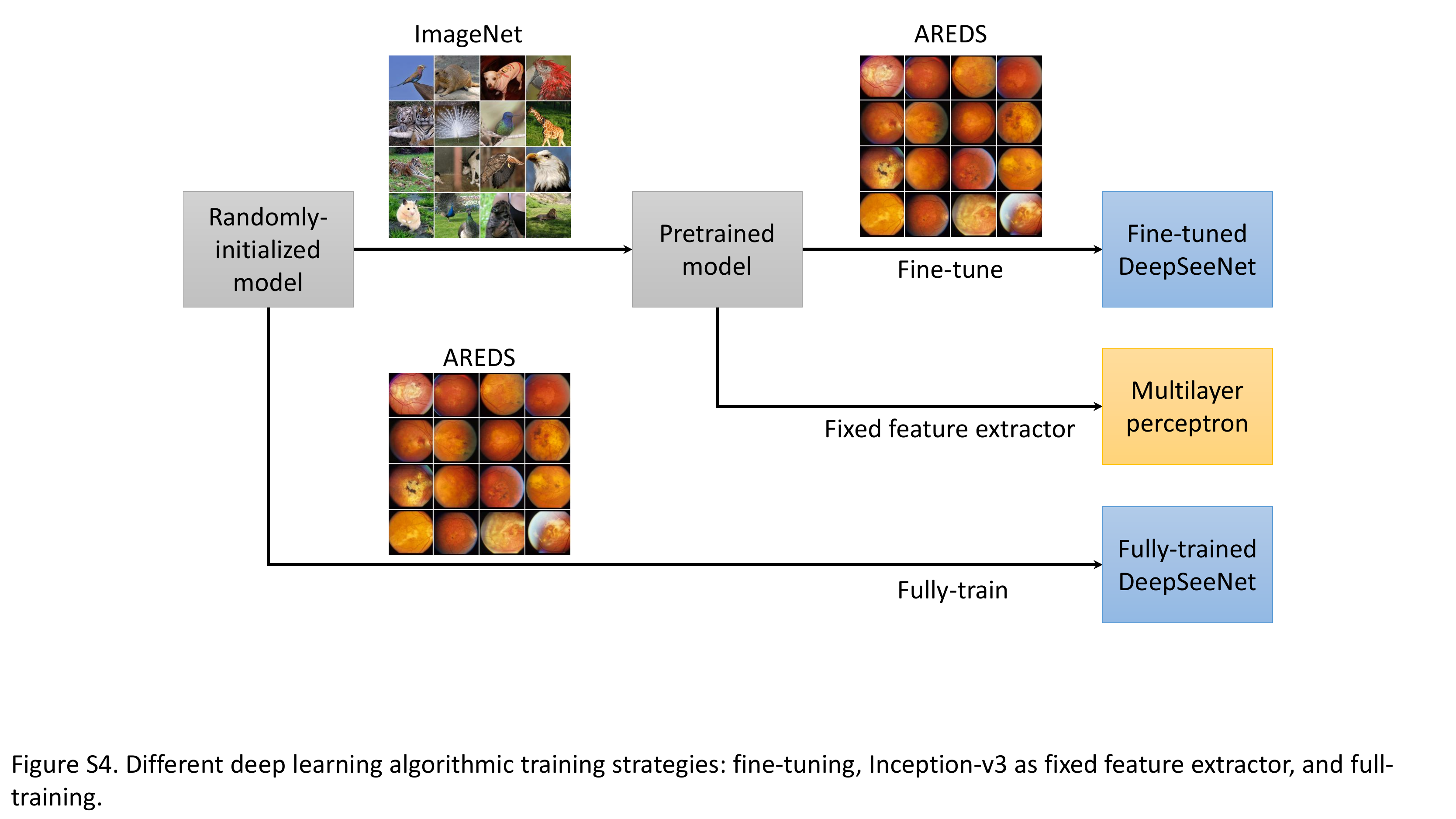}}
    \caption{Different deep learning algorithmic training strategies: fine-tuning, Inception-v3 as fixed feature extractor, and full-training.}
    \label{figs:Different deep learning algorithmic}
\end{figure}

Three training strategies were investigated for model training (Figure~\ref{figs:Different deep learning algorithmic}). (1) We created a `Multi-layer perceptron' (`MLP') model by taking an pre-trained Inception-v3 model on ImageNet~\cite{deng2009imagenet} (an image database of $>14$ million natural images with corresponding labels; see Glossary), then training it on the AREDS dataset as a multi-layer perceptron model (with two densely-connected layers of 256 and 128 units, respectively). In this model, the AREDS images were used to train only the last two layers of the entire model. (2) We created the `Fine-tuned DeepSeeNet' model by taking an pre-trained Inception-v3 model on ImageNet (i.e., same process and same weights as for the MLP model), then training it on the AREDS images. However, for this model, the training process on the AREDS images was used to fine-tune all layers (not just the last two layers). Hence, ultimately, the entire model is trained using the AREDS dataset. (3) Lastly, we created the `Fully-trained DeepSeeNet' model. In this case, we proceeded directly from an Inception-v3 model (with randomly-initialized weights) to training (all layers) using the AREDS images. Hence, in this model, the ImageNet dataset was not utilized for pre-training. More information on the training methodologies can be found in the Supplementary material.
The three training strategies were assessed in turn on the testing dataset: Fine-tuned DeepSeeNet, Fully-trained DeepSeeNet, and MLP. The performance of the three training models is shown in Table~\ref{tabs:Performance of three different deep learning models}. Of the three, the model with the best performance was Fine-tuned DeepSeeNet (i.e., where the model was pre-trained using ImageNet, and all layers were fine-tuned using the AREDS training dataset), with accuracy=0.671 and kappa=0.558.

First, Fine-tuned DeepSeeNet performed more accurately than Fully-trained DeepSeeNet. Fully-trained DeepSeeNet was trained using the same AREDS images, but (unlike Fine-tuned DeepSeeNet) was not pre-trained using the ImageNet images. Hence, Fine-Tuned DeepSeeNet had the advantage of pre-initialized weights. The superior accuracy of this training approach suggests that, although it is possible to train a deep learning model from scratch by using a large new dataset, it is still beneficial in practice to initialize with weights from a pre-trained model (particularly when we consider that there are over 14 million images in ImageNet, compared to 60,000 in the AREDS training set). One possible reason for the inferior accuracy of the model fully trained using the AREDS images is over-fitting, i.e., the model starts to memorize rather than learn to generalize from the training data. A negative consequence is, therefore, that the model has poor performance on the validation dataset~\cite{yosinski2015understanding,yosinski2014how}.

Fine-tuned DeepSeeNet also performed more accurately than the MLP deep learning model. Both deep learning models were pre-trained using the ImageNet images and then trained using the AREDS fundus images. However, the training of Fine-tuned DeepSeeNet was designed specifically to tackle AMD classification: during training using the AREDS images, the weights of all its layers were permitted to change. By contrast, the MLP model underwent training using the same AREDS images, but only as a fixed feature extractor, i.e., the weights of all but two of its layers were not permitted to change. For Fine-tuned DeepSeeNet, we permitted all layers to be retrained using the AREDS dataset, as we considered that the AREDS dataset of images was sufficiently large, with features different from the ImageNet images, to justify optimization of all layers.

\pagebreak

\begin{figure}[!ht]
    \centering
    \frame{\includegraphics[clip,trim=5cm 2cm 5cm 0,width=.8\textwidth]{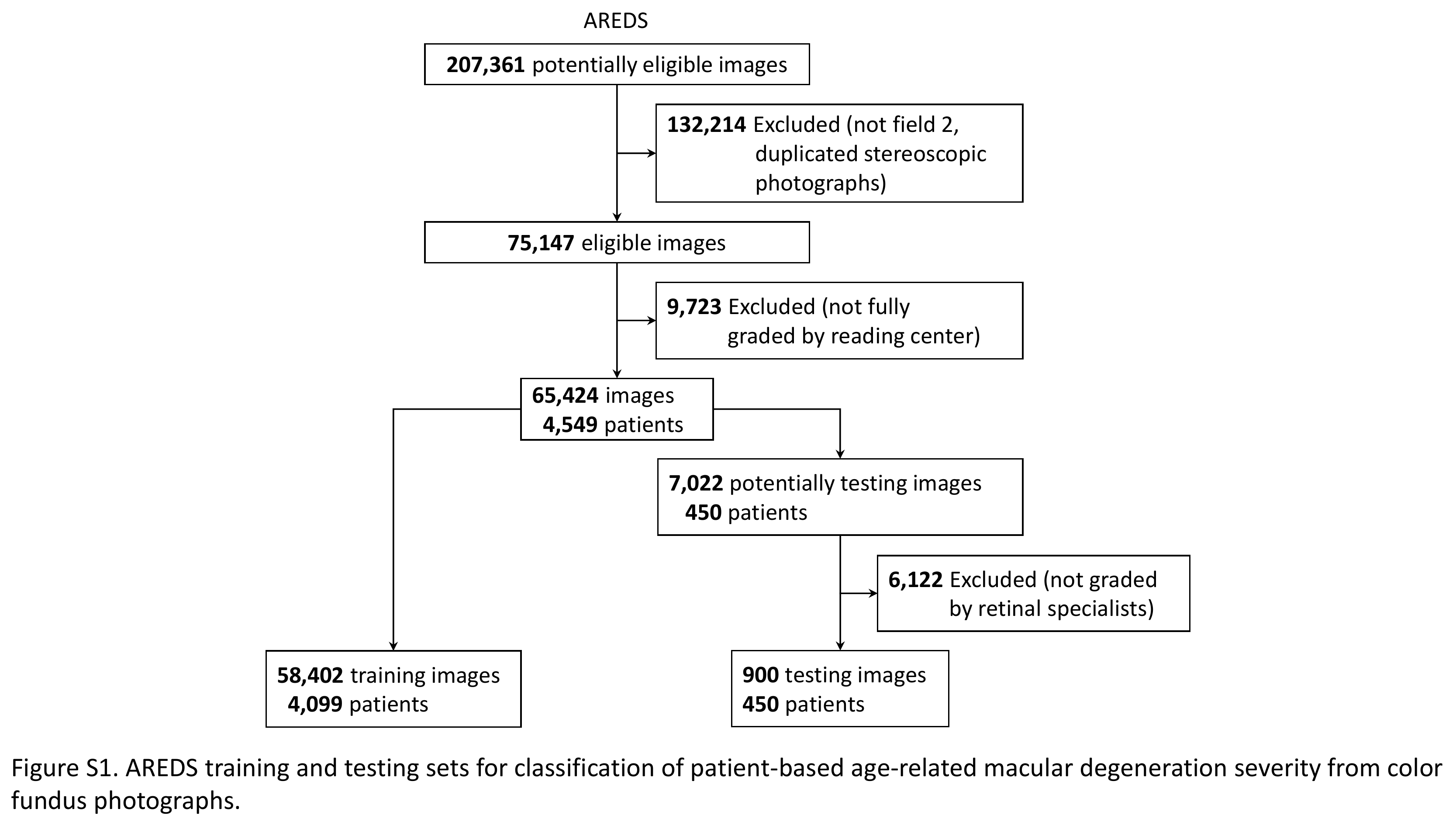}}
    \caption{AREDS training and testing sets for classification of patient-based age-related macular degeneration severity from color fundus photographs.}
    \label{figs:training testing sets}
\end{figure}

\pagebreak

\begin{figure}[!ht]
    \centering
    \frame{\includegraphics[clip,trim=2cm 3cm 1cm 0,width=\textwidth]{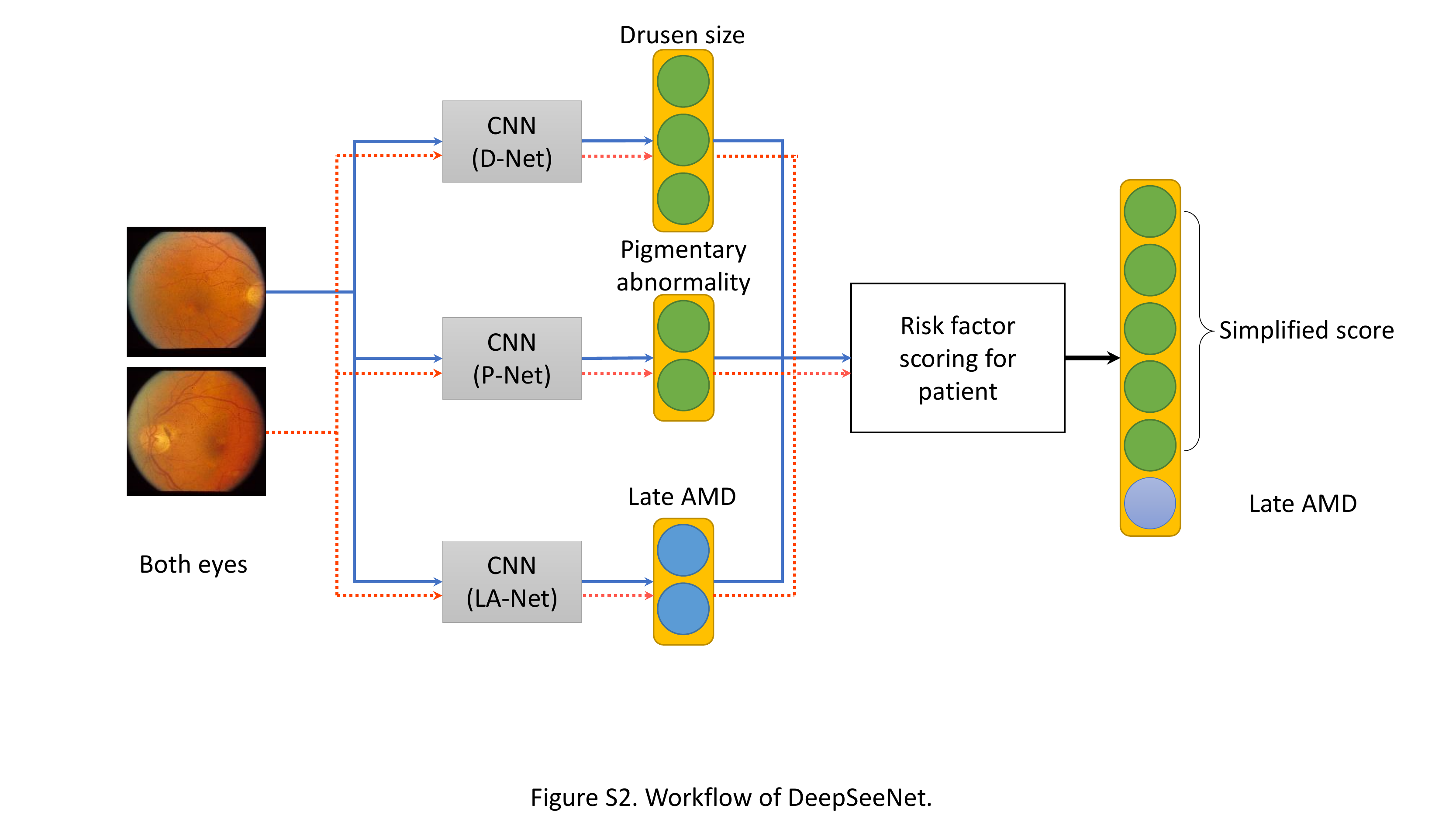}}
    \caption{Workflow of DeepSeeNet.}
    \label{figs:workflow}
\end{figure}

\pagebreak

\begin{figure}[!ht]
    \centering
    \frame{\includegraphics[clip,trim=2cm 4cm 2cm 1cm,width=\textwidth]{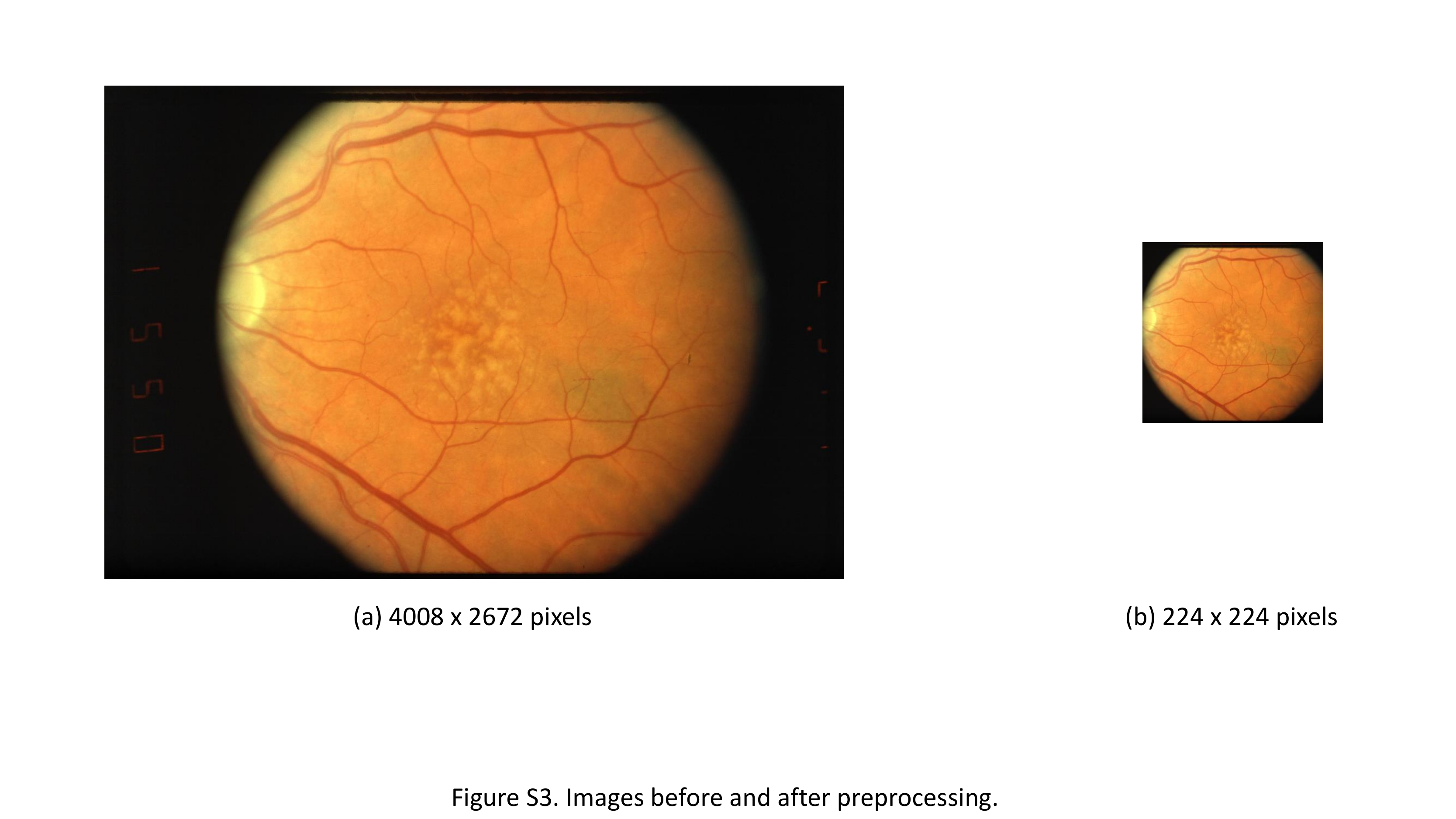}}
    \caption{Images before and after preprocessing.}
    \label{figs:preprocessing}
\end{figure}

\pagebreak

\pagebreak

\begin{table}[!ht]
\caption{Accuracy comparing retinal specialists' performance with that of Fine-tuned DeepSeeNet based on the test set values.}
\label{tabs:Accuracy comparing retinal}
\begin{center}
\begin{tabular}{c@{\hspace{4em}}c@{\hspace{4em}}c@{\hspace{4em}}cr}
    \toprule
    Class & Instance & Retinal specialist & \multicolumn{2}{c}{Fine-tuned DeepSeeNet}\\
    \midrule
    0 & 185 & 83.8\% & 90.8\% &    7.0\\
    1 & 79 & 29.1\% & 43.0\% &  13.9\\
    2 & 56 & 39.3\% & 48.2\% &    8.9\\
    3 & 46 & 26.1\% & 34.8\% &    8.7\\
    4 & 33 & 45.5\% & 72.7\% &  27.2\\
    5 & 51 & 82.4\% & 64.7\% & -17.7\\
    \bottomrule
\end{tabular}
\end{center}
\end{table}

\pagebreak

\begin{table}[!ht]
\centering
\begin{threeparttable}
\caption{Performance of DeepSeeNet on assessing AREDS Simplified Severity Scale scores from images at the study baseline and overall data.}
\label{tabs:Performance of DeepSeeNet on assessing}
\begin{tabular}{l@{\hspace{4em}}cc@{\hspace{4em}}c}
    \toprule
     & Study baseline && Overall\\
    \cmidrule{2-2}\cmidrule{4-4}
     & \multicolumn{1}{r}{(95\% CI)} && \multicolumn{1}{r}{(95\% CI)}\\
    \midrule
    Overall accuracy & 0.671 (0.670-0.672) && 0.662 (0.662-0.623)\\
    Sensitivity      & 0.590 (0.589-0.591) && 0.592 (0.591-0.593)\\
    Specificity      & 0.930 (0.930-0.930) && 0.928 (0.928-0.928)\\
    Kappa            & 0.558 (0.557-0.560) && 0.555 (0.554-0.557)\\
    \bottomrule
\end{tabular}
\begin{tablenotes}
\item CI = confidence interval.
\end{tablenotes}
\end{threeparttable}
\end{table}

\pagebreak

\begin{table}[!ht]
\caption{Performance of three different deep learning models (generated by three different training strategies); on classifying AREDS Simplified Severity Scale scores from color fundus photographs.}
\label{tabs:Performance of three different deep learning models}
\begin{center}
\begin{tabular}{l@{\hspace{2em}}rrrrr}
    \toprule
     & \multicolumn{1}{c}{MLP} && \multicolumn{1}{c}{Fine-tuned DeepSeeNet} && \multicolumn{1}{c}{Fully-trained DeepSeeNet}\\
    \cmidrule(r){2-2}\cmidrule(r){4-4}\cmidrule{6-6}
    & (95\% CI) && (95\% CI) && (95\% CI)\\
    \midrule
    Overall accuracy & 0.436 (0.435-0.437) && 0.671 (0.670-0.672) && 0.624 (0.623-0.625)\\
    Sensitivity      & 0.236 (0.235-0.236) && 0.590 (0.589-0.591) && 0.494 (0.493-0.495)\\
    Specificity      & 0.862 (0.862-0.862) && 0.930 (0.930-0.930) && 0.919 (0.918-0.919)\\
    Kappa            & 0.163 (0.162-0.164) && 0.558 (0.557-0.560) && 0.487 (0.486-0.488)\\
    \bottomrule
\end{tabular}
\end{center}
MLP = multilayer perceptron; CI = confidence interval.
\end{table}

\end{document}